 \documentclass[preprint,review,12pt]{elsarticle}

\usepackage{float}

\usepackage{amsmath,amssymb,amsfonts}
\usepackage{algorithmic}
\usepackage{algorithm2e}
\usepackage{textcomp}
\usepackage{float}
\usepackage{longtable}

\usepackage{amssymb}

\journal{Computer Methods and Programs in Biomedicine}

\begin{document}

\begin{frontmatter}

\title{AI Driven Knowledge Extraction from Clinical Practice Guidelines: Turning Research into Practice}






\author{Musarrat Hussain}
\ead{musarrat.hussain@oslab.khu.ac.kr}

\author{Jamil Hussain}
\ead{jamil@oslab.khu.ac.kr}

\author{Taqdir Ali}
\ead{taqdir.ali@oslab.khu.ac.kr}

\author{Fahad Ahmed Satti}
\ead{fahad.satti@oslab.khu.ac.kr}

\author{Sungyoung Lee\corref{cor1}}
\cortext[cor1]{Corresponding author}
\ead{sylee@oslab.khu.ac.kr}

\address{Department of Computer Science and Engineering, Kyung Hee University, South Korea}

\begin{abstract}
\textbf{Background and Objectives:} Clinical Practice Guidelines (CPGs) represent the foremost methodology for sharing state-of-the-art research findings in the healthcare domain with medical practitioners to limit practice variations, reduce clinical cost, improve the quality of care, and provide evidence based treatment. However, extracting relevant knowledge from the plethora of CPGs is not feasible for already burdened healthcare professionals, leading to large gaps between clinical findings and real practices. It is therefore imperative that state-of-the-art Computing research, especially machine learning is used to provide artificial intelligence based solution for extracting the knowledge from CPGs and reducing the gap between healthcare research/guidelines and practice. \\
\textbf{Methods:} This research presents a novel methodology for knowledge extraction from CPGs to reduce the gap and turn the latest research findings into clinical practice. First, our system classifies the CPG sentences into four classes such as condition-action (C-A), condition-consequences (C-C), action (A), and not-applicable (NA) based on the information presented in a sentence. We use deep learning with state-of-the-art word embedding, improved word vectors technique in classification process. Second, it identifies qualifier terms in the classified sentences, which assist in recognizing the condition and action phrases in a sentence. Finally, the condition and action phrase are processed and transformed into plain rule \textit{\textbf{If} Condition(s) \textbf{Then} Action} format. \\
\textbf{Results:} We evaluate the methodology on three different domains guidelines including Hypertension, Rhinosinusitis, and Asthma. The deep learning model classifies the CPG sentences with an accuracy of 95\%. While rule extraction was validated by user-centric approach, which achieved a Jaccard coefficient of 0.6, 0.7, and 0.4 with three human experts extracted rules, respectively. \\
\textbf{Conclusions:} The methodology is beneficial in transforming the latest research findings into an executable format that can be directly unitized by healthcare systems to assist in making the right clinical decisions.
\end{abstract}


\end{frontmatter}



\section{Introduction}

Text mining has gained substantial attention and becomes extremely important due to the increase in the number of online textual resources \cite{kluegl2016uima}. Text mining is a broader term used for assisting retrieval and manipulation of text documents for useful information extraction and has revealed a variety of applications in various domains \cite{pandita2017tmap}. Like other domains, the clinical domain is also influenced by the information explosion. Various types of text documents, including clinical research papers, protocols, and clinical practice guidelines (CPGs) are published online. These documents contain valuable information and have a potential to assist in  improving healthcare services.     

CPGs are one of the valuable resources that need to be considered for extracting beneficial information to restrain inappropriate healthcare service provision. CPG is defined as: ``Clinical Practice Guidelines are statements that include recommendations intended to optimize patient care that is informed by a systematic review of evidence and an assessment of the benefits and harms of alternative care options" \cite{pyon2013primer}. It provides evidence based scientific guidelines to healthcare service providers and patients for making appropriate decisions for a given medical condition \cite{heins2016adherence}. CPGs are comprehensive enough that covers all possible interventions appropriate for a particular disease \cite{ruan2015clinical}.  Thousands of CPGs have developed annually for the improvement of healthcare quality and scientific evidence-based treatment \cite{upshur2014clinical}. The developed CPGs are published online in textual format, which is inadequate for the healthcare service providers to access, understand, evaluate, and apply it in a limited time during real practices. This leads to a gap between the latest research findings and clinical practices \cite{ryan2017adherence}. The medical research is only beneficial if it is applicable in real practice \cite{pyon2013primer}.  

The researchers have applied various text mining techniques to automate some of the steps in translating the research findings into practice and reducing its gap. Mostly, the researchers focused medical concepts and temporal relationship extraction, entities and event identification, and their relationship identification \cite{wang2018clinical, sun2013evaluating, sarawagi2008information, small2014review}. These techniques are beneficial but not enough to deal with patient specific situations. Therefore, there is an imperative need of a methodology that can identify and extract disease oriented knowledge from the CPG and assist healthcare providers to standardize healthcare services.

This research mainly focuses on automatic knowledge extraction from CPGs in the form of plain rules \textit{\textbf{IF} Condition(s) \textbf{THEN} Action}. To achieve this goal, we classify CPG sentences into four categories, including condition-action (C-A), condition-consequences (C-C), action (A), and not-applicable (NA) based on the information presented in each sentence. We train a deep learning model on experts annotated CPGs that can classify the unseen CPG sentences into one of the aforementioned categories. The model consists of four major layers, including embedding, deep network, fully connected, and output layer. In the embedding layer, we use Improved Word Vector (IWV) for generating the word vectors. Likewise the IWV study \cite{rezaeinia2019sentiment}, we use the combination of Word2Vec, POS2Vec, and Word-position2Vec for creating word vectors. The deep network layer transforms the embedding vectors to compressed representation, which captures the information presented in the sequence of words in the CPG text. The fully connected layer transforms the compressed representation into the output class score while the output layer assign final output based on the class scores using softmax function \cite{rezaeinia2019sentiment}. 

The sentences of each category contain qualifier terms that can assist in identification of condition and action phrases, where qualifiers are general terms that enhance or reduce the intensity or meaning of another word \cite{hettne2010rewriting}. Our proposed system extracts qualifiers by considering its importance and relevancy to the sentence category using weighting criteria such as weight by correlation, weight by gini index, weight by information gain, and weight by information gain ration . The weights assigned to each qualifier is aggregated to get single weight values against each qualifier. The qualifiers having weight less than $\alpha$ are filtered out, and a final list of the qualifiers is achieved. The system also uses word expansion techniques using pre-trained word embedding models to get an enhanced list of the qualifiers. The system thoroughly examines the extracted qualifiers to identify the condition's and action's direction information with respect to qualifier. For instance a qualifier with condition's direction \textit{LEFT} indicates that the conditioning phrase lies to the left side of the qualifier in the CPG sentence. We tokenize condition phrases and find the semantic category of each token using unified medical languages system (UMLS) dictionary. The terms having semantic categories related to a condition are listed as condition terms. However, we required condition term, operator, and  value to represent a complete condition. Therefore, we evaluate neighbor terms for operators and term values. Similarly, we follow the same procedure for action term identification. Finally, we combine all identified conditions and corresponding actions to the plain rule format. 

The proposed methodology automates the knowledge extraction from CPGs, that can directly be utilized by healthcare service providers or can be used to assist in healthcare decisions making. It reduces the gap between research and practice by minimizing the human efforts required for CPGs evaluation, analysis, and understanding.

The rest of the study is organized as follows. Section \ref{RelatedWork} presents related work, and Section \ref{ProposedMethodology} describes the details of the steps followed for CPGs sentence classification and techniques used for knowledge extraction from the sentences. Section \ref{ResultsEvaluation} presents the results obtained by the sentence classification and rule generation. Finally, Section \ref{Conclusion} concludes the study with a summary of the research findings and future directions.

\section{Related Work}
\label{RelatedWork}

The National Institutes of Health Consensus Development Program commenced CPGs development in late 1970. The objective of the program was to determine and stimulate the best practice for the improvement of healthcare quality \cite{jacobsen2009clinical}. To convert the CPG into computer interpretable form, The clinical information needs to be extracted and represented in a computable form.  A tremendous research work has been done to extract clinical information and represent it into various computable model. Most of the researchers have processed it for specific information of their interest. Such as Shiyi Zhao et al. \cite{zhao2019associative}  have proposed an associative attention networks for extraction temporal relation extraction from Electronic Health Records (EHR). Youngjun Kim et al. \cite{kim2019ensemble} proposed an ensemble method for the extraction of medication and relation information from clinical text. Focil-Arias et al. \cite{viegas2019cluwords} have processed clinical records for medical event extraction using conditional random fields. Engy Yehia et al. \cite{viegas2019cluwords} proposed OB-CIE, an ontology based clinical information extraction system. They processed physician's free-text notes for medical concepts and represents them in the ontological form. The ontology model is used to assist physicians in documenting visit notes without changing their workflow. Shuai Zheng et al. \cite{zheng2017effective} proposed clinical information extraction framework called IDEAL-X, which works on the top of online machine learning. While processing a document it records user interaction with the system as feedback and update the learning model accordingly.

The Informatics for Integrating Biology and the Bedside (i2b2) \cite{i2b2} has designed various clinical text processing tasks and challenges such as obesity Challenge where the participated research need to identify either the patient is obese and what co-morbidities do they have from their clinical textual records. Similarly, medication extraction challenge, relations challenge, temporal relations challenge, and heart disease challenge was design to process and extract concern information from clinical text \cite{i2b2Challenges}. Also, National NLP Clinical Challenges (n2c2) has shared task on challenges in processing for clinical data \cite{n2c2}. Wendy W Chapman et al. \cite{chapman2011overcoming} have described the rule of shared task in overcoming the barriers in clinical text processing.  A number of comprehensive surveys and comparison among various guidelines modeling schemes have been presented in  \cite{wang2018clinical, kreimeyer2017natural, alves2019information, sun2018data, zhang2019deep, peleg2003comparing, bottrighi2009analysis, ten2008computer}.

Various guideline document models and tools have also been introduced to represent the textual CPG in computer understandable format \cite{latoszek2010clinical, peleg2003comparing, qiu2018automated, khodambashi2017reviewing, de2004approaches}. The common approaches which are used in the guidelines representations includes but not limited to Asbru \cite{shahar1998asgaard}, Guideline Element Model (GEM) \cite{shiffman2000gem}, GuideLine Interchange Format GLIF \cite{ohno1998guideline}, Arden Syntax \cite{kim2008modeling}, PROforma \cite{sutton2003syntax}, and SAGE \cite{tu2007sage}.  Asbur \cite{shahar1998asgaard} assist in CPG annotation and task oriented CPG ontological modeling. GEM \cite{shiffman2000gem} is a widely used XML based computable model, uses more than a hundred multilevel hierarchical tags to classify and represent a CPG information. The user determines text from a CPG, perform necessary modifications like filter stop words, and converting passive to active voice, and assign it to one of the appropriate tags in the GEM model XML file. GLIF \cite{ohno1998guideline} represent the CPG model for sharing guidelines knowledge among various healthcare organizations. These models represents the CPG knowledge into computer interpretable format that can be used by physicians in utilizing CPG in real practices.

The aforementioned approaches of clinical information extraction and modeling  are useful for CPG processing, but they targeted limited information like diseases, drug events, and temporal information. Also, it required extensive human involvement, which causes difficulties in CPG processing and modeling, and limits their adoption in clinical workflows. Therefore, there is an utmost need for an automated technique that can extract complete knowledge and transform the textual CPG to computer interpretable format with less or no human involvement. The human involvement should be limited to model verification instead of manual CPG encoding. Our proposed system focuses on encoding the CPG to a simple and effective form (\textbf{\textit{IF}} conditions \textbf{\textit{THEN}} action) using state-of-the-art machine learning techniques.         

\section{Proposed Methodology}
\label{ProposedMethodology}

\begin{figure}[H]
	\centering
	\includegraphics [scale=0.83]{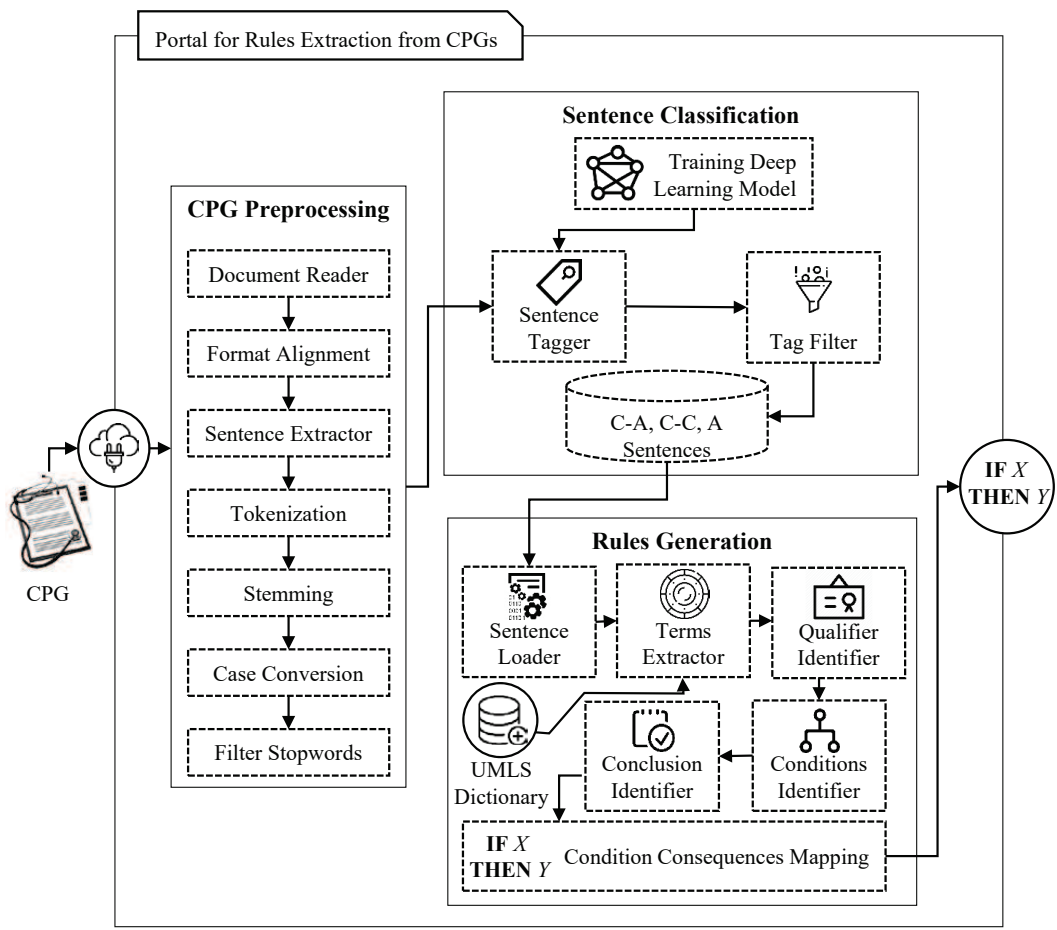}
	\caption{Proposed methodology workflow.}
	\label{fig:methodology}
\end{figure}

In this research, we present an end-to-end methodology for automatically transforming CPGs text into a computer understandable format such as plain rules. The major steps involved in achieving this goal include CPG preprocessing, sentence classification, and rules generation as shown in Figure \ref{fig:methodology}. Where CPG preprocessing splits the content of the CPG into sentences and performs a necessary text processing steps, including word tokenization, stemming, case conversion, and stop words filtration. Sentences classification categorizes the sentences to one of the four categories condition-action (C-A), condition-consequences (C-C), action (A), and not applicable (NA) based on the characteristics and terms used in a sentence. While rules generation scrutinizes the classified sentences to find conditions and their corresponding action parts and transforms into plain rules format. The details of each step are described in the following subsections.

\subsection{CPG Preprocessing}
\label{CPGPreprocessing}     

The preprocessing like wise other applications, aligns the CPG format to analyzable form according to requirements. The building blocks required for our methodology is sentences and word tokens, therefore, the preprocessing split CPG contents into individual sentences and further to tokens. Some preliminary preprocessing tasks like tokenization, stemming, case conversion, and filter stop words has been performed. Sentences and the extracted tokens are passed to the subsequent components for further processing.

\subsection{Sentence Classification}
\label{SentenceClassification}

The importance of CPG sentences depends on the presented information and targeted goal. Some sentences may not be important as others for a target application. Therefore, we need to filtered out the important sentence only. In this study, we used an existing dataset of the annotated guidelines, Hypertension \cite{james20142014}, Rhinosinusitis \cite{chow2012idsa}, and chapter four of  Asthma guideline \cite{british2003scottish}. The CPG sentences are categorized by domain experts into four classes such as C-A, C-C, A, and NA based on the information present in a sentence. C-A sentences explain about actions required in particular situations or conditions. C-C describes the consequences of particular conditions. Some sentences only explain what sort of action is required, represented by `A'. The condition for these sentences can be extracted from the context or from the neighbor sentences, while NA represent the background information or thought of the authors. These sentences are not of interest, generally.     

\begin{figure}[H]
	\centering
	\includegraphics [scale=0.63]{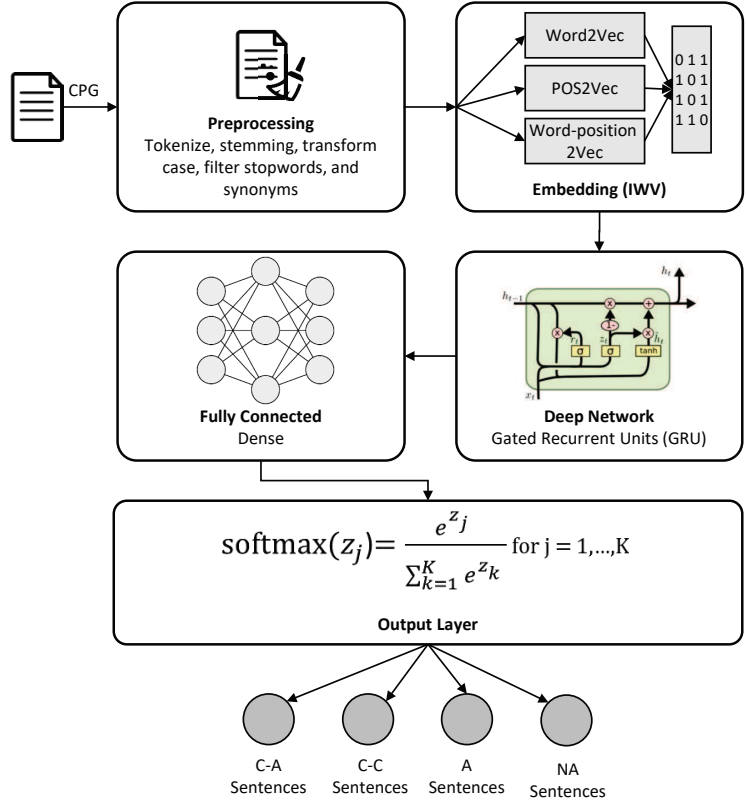}
	\caption{Deep learning model for CPG sentence classification.}
	\label{fig:sentenceClassifier}
\end{figure}

Our primary focus is to extract knowledge in the form of plain rules from CPGs. Therefore, our target of interest sentences includes C-A, C-C, and A. Manual sentence classification is a cumbersome task and required time, human resources with detailed domain knowledge as well as CPG understanding. Therefore, we train a deep learning model that can automatically classify CPG sentences into classes C-A, C-C, and A. The model architecture consists of four major layers, embedding (IWV), deep network, fully connected, and output layer. The goal of the embedding layer is to represent the CPG text into a vector format. Mostly, the researchers used the highly accurate techniques such as Word2Vec and GloVe \cite{mikolov2013efficient}, \cite{pennington2014glove} in their studies. Therefore, we convert CPG text into vector using improved Word2Vec technique IWV in embedding layer \cite{rezaeinia2019sentiment}. In our model, we utilize the POS tagging technique, word position algorithm, and Word2Vec to generate final vector representation for the embedding layer. Deep network layer processes the embedding layer's output and captures information, presented in the sequence of the word. In this layer, we used gated recurrent unit (GRU) for efficient results. The fully connected layer identifies the final classes score calculation. While, the output layer predicts the final classes using softmax function. The trained model is then able to classify unseen CPG sentences to their appropriate categories efficiently. The classified sentences help in condition and action phrases identification, which are used in subsequent components to acquire the associated knowledge. The detail of the classification model is shown in Figure \ref{fig:sentenceClassifier}.

\subsection{Rules Generation}
\label{RulesGeneration}

\begin{figure}[H]
	\centering
	\includegraphics [scale=0.52]{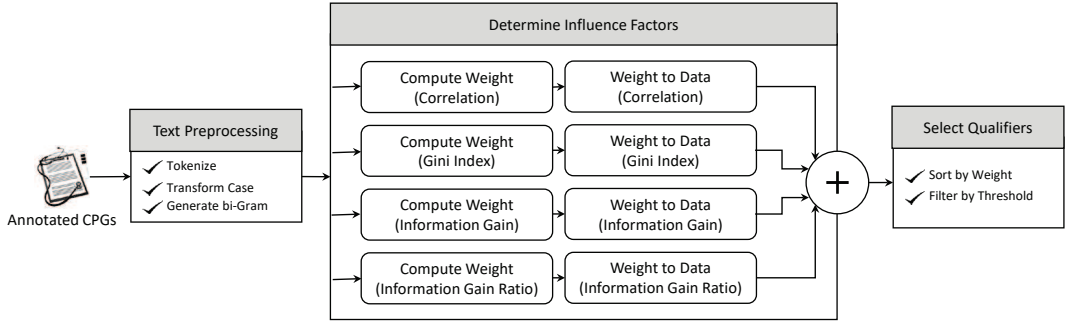}
	\caption{Qualifiers extraction process.}
	\label{fig:qualifierExtractionProcess}
\end{figure}

Rules generation is the most critical and complex part of the study. It requires an in-depth analysis of each classified sentence. The primary hindrance in rule generation is the  recognition of condition(s) and associated action(s) phrases in a sentence. However, there exist some clue words known as a qualifier in most of the sentences that can assist the process. Therefore, the system identifies and extracts a list of qualifiers from annotated CPG, automatically as depicted in Figure \ref{fig:qualifierExtractionProcess}. The system identifies and extracts qualifiers from CPGs in three sequential steps. As a first step, the system performs CPGs essential preprocessing. However, in preprocessing, we did not remove stop words from CPG contents, because, it clarifies the importance of qualifier as well as assists in identifying direction information. Also, the individual tokens were not useful as the combination of multiple tokens. Therefore, we use bi-gram while generating tokens to be used in the subsequent process. In the second step, the system identifies  a list of qualifiers. The qualifier terms were determined using various weights techniques including correlation, gini index, information gain, and gain ratio, as our previous work \cite{hussain2018multimodal}. The weights assigned to each term are aggregated to get the final weight of each term. Finally, the terms sorted by weight values, and a filter of weight greater than fifty was applied to get a list of most import qualifiers $Q$. A list of fifty qualifiers were extracted and after applying filter, total twenty qualifiers were having weight greater than fifty as shown in Table \ref{tab:qualifierwithWeights}.

\begin{table}[H]
	\caption{List of twenty qualifiers with their weights}
	\setlength{\tabcolsep}{3pt}
	\centering
	\begin{tabular}{|l|l|l|l|l|l|}
		\hline
		S. No & Qualifier &  Weight & S. No & Qualifier &  Weight\\
		\hline
		1 & leads to & 235 & 11 & to lower & 107   \\
		2 & treatment with & 194 & 12 & applies to & 106  \\
		3 & in population  & 182 & 13 & panel decide & 99  \\ 
		4 & general treatment & 167 & 14 & patient with & 97  \\
		5 & recommended for & 145 & 15 & goal should & 89  \\
		6 & with disease & 145 & 16 & for outcome & 75 \\
		7 & is needed & 133 & 17 & start drug & 72 \\
		8 & in black & 115& 18 & in opinion & 67  \\
		9 & initiate the & 115 & 19 & use of & 53  \\
		10 & initiate the & 111 & 20 & improvements in & 51  \\
		
		\hline
	\end{tabular}
	\label{tab:qualifierwithWeights}
\end{table}

The final qualifier list may not perform well on other domains because of lack of generalization. Therefore, we remove this deficiency using the word expansion technique by utilizing pre-trained word embedding models as discussed in \cite{viegas2019cluwords}. The $Q$ is the list of extracted qualifiers presented in a set of CPG documents $D$ and $V$ is the set of vectors representing each qualifier in $Q$ according to pre-trained word embedding (Fasttext and Word2Vec). Therefore, each qualifier $q$ in $Q$ corresponds to a vector $v$ $\in$ $V$ having length $l$, where $l$ represents word vector space dimensionality. We outline the extended candidate qualifiers as a matrix $C \in \mathbb{R}^{|Q| \times |Q|}$, where each index $C_{i, i^\prime}$ is calculated using Equation \ref{eqn:qualifierIndex}.

\begin{equation}
\label{eqn:qualifierIndex}
C_{i, i^\prime} = 
\begin{cases}
\omega(i, i^\prime) & \text{if}\ \omega(i, i^\prime) \geq \alpha \\
0, & \text{otherwise}
\end{cases}
\end{equation}

Where $\omega(i, i^\prime)$ represents the cosine similarity  between term $i$ and $i^\prime$, and $\alpha$ represents the threshold values which is used to filter irrelevant terms having cosine similarity less than $\alpha$. i.e $\omega(i, i^\prime) < \alpha $. Here each candidate qualifier is shown as a row $C_i$, and each column $i^\prime$ in $Q$ corresponds to the component in $C_i$.  The cosine similarity between two terms $u$ and $v$ is calculated using Equation \ref{eqn:cosineSimilarity}. The cosine similarity between terms lies in the range zero to one where $\alpha = 0$ means the two terms are entirely different while  $\alpha = 1$ means both terms are exactly similar. We filter out all terms having similarity values less then 0.5 i.e $\alpha > 0.5$ and the remaining terms are considered as qualifiers. The final enriched set of the  qualifier is used in the following steps for condition and action identification and rules generation. An example of the qualifier ``recommend" expansion is shown in Table \ref{tab:qualifierExpansion}. We used the English Wikipedia and GoogleNews pre-trained models for word expansion.

\begin{table}[H]
	\caption{Example of the words similar to ``recommend"}
	\setlength{\tabcolsep}{1pt}
	\centering
	\renewcommand{\arraystretch}{1}
	\begin{tabular}{|l|l|}
		\hline
		\multicolumn{2}{|c|}{recommend} \\
		\hline
		Semantically similar words & advise, request, caution, urge, approve, \\ & prescribe, advocate, encourage, instruct, suggest    \\
		& like, choose, wanted, decided, need, find, consider, \\
		& considered, propose, consult, urge, concur, approve, \\
		& prefer, endorse, decide \\
		Syntactically similar words & recommended, recommending, recommends,\\ &recommendation, recommendations, commend   \\
		\hline
	\end{tabular}
	\label{tab:qualifierExpansion}
\end{table}

\begin{equation}
\label{eqn:cosineSimilarity}
\omega(u, v) = \frac{\sum_{j}^{l}u_j.v_j}
{\sqrt{\sum_{j}^{l}u_j^2}.\sqrt{\sum_{j}^{l}v_j^2}}
\end{equation}

We exhaustively analyze the CPG sentences of each category for  qualifier identification. Qualifier of a sentence can also direct us about condition phrase and action phrase location in a sentence. The example subset of the identified qualifiers, along with direction of condition and action, are given in Table \ref{tab:qualifierList}. Where the condition direction shows the possible position of the condition phrase and action direction represents the action phrase possible direction with respect to the qualifier. For example, in the CPG sentence that has qualifier ``leads to",  the condition phrase is usually located before the qualifier while action phrase after the qualifier. Therefore, the condition direction of the qualifier is set to \textit{Left} and action  direction to \textit{Right}.  

\begin{table}[H]
	\caption{List of qualifiers with condition and action directions}
	\label{table}
	\setlength{\tabcolsep}{3pt}
	\begin{tabular}{|p{30pt}|p{120pt}|p{100pt}|p{100pt}|}
		\hline
		S. No & Qualifier & Condition Direction & Action Direction \\
		\hline
		1 & leads to 		& Left  & Right \\
		2 & recommended for & Right & Left  \\
		3 & applies to  & Right & Left \\ 
		4 & treatment with & Right & Right \\
		5 & is recommended & Right & Right \\
		6 & in ... initiate .... & Left & Right \\
		7 & ... is needed to ...  & Right & Left \\
		8 & in ... goal should be .... & Left & Right \\
		9 & .... improvements in ... & Left & Right \\
		10 & .. use of ... in... & Right & Right \\
		
		\hline
	\end{tabular}
	\label{tab:qualifierList}
\end{table}

To identify conditions and corresponding action for rules generation, we split each sentence into two parts on the bases of qualifier , left phrase $SP_{left}$ and right phrase $SP_{right}$. The qualifiers are having condition  and action directions on opposite direction such as one to the left and other to the right can easily be identified base on the direction. However, some qualifiers  have condition and action on the same side, which increases the complexity of separating condition and action phrases. For example, in the sentence ``In the black hypertensive population , including those with diabetes , a calcium channel blocker or thiazide-type diuretic is recommended as initial therapy" the condition and action phrases lies to the left of the qualifier ``is recommended".  In this type of cases, we consider comma (,)  for  separating the action and condition phrases. The last comma present at the qualifier direction distinguish condition phrase from action.

\begin{figure}[H]
	\centering
	\includegraphics [scale=0.53]{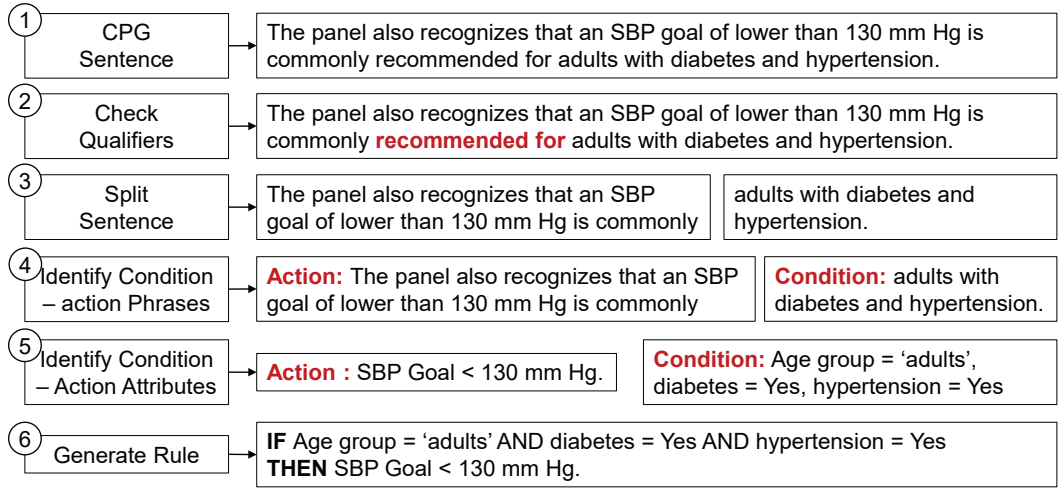}
	\caption{Rule generation example.}
	\label{fig:ruleExample}
\end{figure}

The identified condition phrases are the sequence of strings that may contain multiple condition terms, their values, and some irrelevant terms. Therefore, it needs further processing to extract only condition relevant terms from the phrase. Considering each phrase may restrain the solution and make it CPG dependent. Therefore, we overcome specialization deficiency and make the solution applicable more generalize using  semantic categories of the UMLS dictionary. We identify two lists of UMLS categories that can be consider as condition and action, respectively. To pick the condition terms: also called Keys, from a condition phrase, we identify each token's semantic type using UMLS and find in the condition list. We find operator $O$ and value $V$ for each key $K$  by considering  its neighbor terms to left and right sides (context window). The context window size was set to two. We symbolize each key, operator, and value into triple $<K,O, V>$ form. For example the triple $<SBP, >, 90>$ represents the condition \textit{\textbf{if} systolic blood pressure $>$ 90}. In similar manner, we process the action phrase to extract the action term. Finally, we represent the extracted triples into plain rule such as \textit{\textbf{If} Conditions \textbf{Then} Action} format. As an example of the CPG, C-A sentence, ``The panel also recognizes that an SBP goal of lower than 130 mm Hg is commonly recommended for adults with diabetes and hypertension" was processed by the methodology and the rule "\textit{\textbf{If} Age Group =  adult AND diabetes = Yes AND hypertension = Yes \textbf{Then} SBP Goal $<$ 130 mm Hg}" was extracted. A stepwise example of the proposed methodology is shown in Figure \ref{fig:ruleExample}. The detailed algorithm of the methodology is given in  Algorithm \ref{alg:ruleGeneration}.

\begin{algorithm}[H]
	\fontsize{8}{11}\selectfont
	\SetAlgoLined
	\SetKwInOut{Input}{Input}\SetKwInOut{Output}{Output}
	\Input{
		Guideline $G$, Qualifiers $Q = \{q_1, q_2, q_3, ..., q_n \}$, ConditionCategories $CC = \{cc_1, cc_2, cc_3, ..., cc_n\}$, 
		ActionCategories $AC = \{ac_1, ac_2, ac_3, ..., ac_n\}$ \\
	}
	\KwResult{RuleSet $R = \{r_1, r_2, r_3, ..., r_n\}$ }
	SentenceList $S \gets SplitDocToSentences(G)$ \\
	\ForEach{$s_i$ in $S$}{
		Tokens $T \gets tokenize(s_i)$ \\ 
		TokensSemanticType $TST \gets []$\\
		\ForEach{$t_j$ in $T$}
		{
			SemanticType $ST_j \gets GetTokenSemanticType(t_j)$ \\
			$TST.push(\{t_j, ST_j\})$ \\
		}
		\ForEach{$q_k$ in $Q$} 
		{
			$Matched \gets Find(q_k, TST)$ \\
			\If{$(Matched)$}{
				SentenceParts $SP \gets Split(s_i, q_k)$ \\
				conditionDirection $CD_{qk} \gets getConditionDirection(q_k)$ \\
				actionDirection $AD_{qk} \gets getActionDirection(q_k)$  \\
				
				\If{($CD_{qk} = LEFT$ \textbf{and} $AD_{qk} = RIGHT$)}{
					$sentenceConditionPhrase \gets SP_{left}$ \\
					$sentenceActionPhrase \gets SP_{right}$ \\
				} \ElseIf{($CD_{qk} = RIGHT$ \textbf{and} $AD_{qk} = LEFT$)}{
					$sentenceConditionPhrase \gets SP_{right}$ \\
					$sentenceActionPhrase \gets SP_{left}$ \\
				} \Else{
					$sentenceConditionPhrase \gets IdnetifyCondtionPhrase(SP)$ \\
					$sentenceActionPhrase \gets IdnetifyActionPhrase(SP)$ \\
				} 
				\ForEach{token $t_l$ in $sentenceConditionPhrase$}{
					\If{$t_l$ existIn $CC$}{
						$Key_l \gets t_l$ \\
						$Operator_l \gets findTokenOperator(t_l)$ (see Algorithm. 2)\\
						$Value_l \gets findTokenValue(t_l)$ \\
					}
				}
				$action \gets findAction(sentenceActionPhrase)$
			}	
		}
	}
	\caption{Rules extraction from clinical practice guidelines}
	\label{alg:ruleGeneration}
\end{algorithm}

\section{Results and Evaluation}
\label{ResultsEvaluation}

The proposed methodology is evaluated on existing three datasets, which  includes Hypertension \cite{james20142014}, Rhinosinusitis \cite{chow2012idsa}, and chapter four of  Asthma guideline \cite{british2003scottish}. The details of the datasets are given in Table \ref{tab:dataset}. The result achieved by each step of the proposed methodology is described in the following subsections. 

\begin{table}[H]
	\caption{Details of used dataset}
	\setlength{\tabcolsep}{3pt}
	\begin{tabular}{|p{80pt}|p{40pt}|p{40pt}|p{40pt}|p{40pt}|p{100pt}|}
		\hline
		Guidelines & C-A & C-A & A & NA & Total Sentences \\
		\hline
		Hypertension & 60  & 14 & 4 & 200 & 278 \\
		Rhinosinusitis & 97 & 39 & 15 & 610 & 761 \\
		Asthma & 38 & 7 & 8 & 119 & 172 \\
		\hline
		Total & 198 & 60 & 24 & 929 & 1211 \\
		
		\hline
	\end{tabular}
	\label{tab:dataset}
\end{table}

\subsection{Sentence Classification }
\label{ResultsSentenceClassification}

\begin{figure}[H]
	\centering
	\includegraphics [scale=0.96]{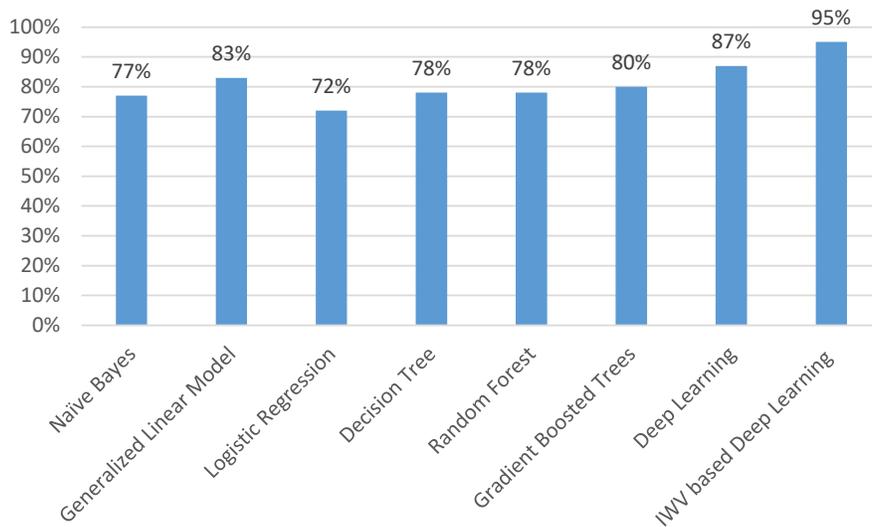}
	\caption{Sentences classification models accuracies.}
	\label{fig:classificationResult}
\end{figure}

In this study, we combined the sentences of all three guidelines Hypertension \cite{james20142014}, Rhinosinusitis \cite{chow2012idsa}, and chapter four of  Asthma guideline \cite{british2003scottish}. We trained various machine learning models, including Naive Bayes, Generalized Linear Model, Logistic Regression, Decision Tree, Random Forest, Gradient Boosted Trees, Deep Learning, and IWV based Deep Learning to check the CPG sentences classification performance in terms of accuracies. We used the 70/30 training, testing ratio of the dataset for models training and evaluation. All the experiments have been carried out in Rapidminer tool \cite{mierswa2006yale} and Tensorflow with  deep learning library \cite{abadi2016tensorflow}.  All reports are generated based on the average accuracies computed over multiple runs of models.  The results achieved by the classification models are 77\%, 83\%, 72\%, 78\%, 78\%, 80\%, 87\%, and 95\%, respectively, as shown in Figure \ref{fig:classificationResult}.

\begin{figure}[H]
	\centering
	\includegraphics [scale=0.90]{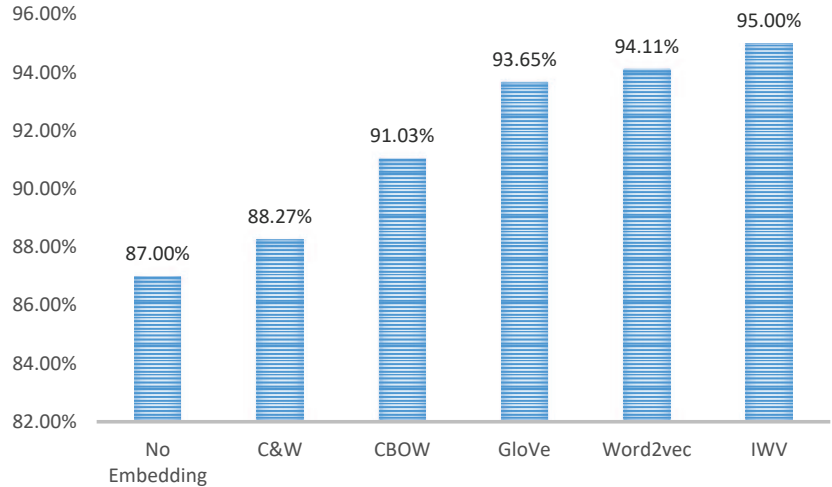}
	\caption{Model Accuracy with various embeddings.}
	\label{fig:wordEmbeddingAccuracy}
\end{figure}

As shown in Figure \ref{fig:classificationResult}, various models produced different accuracies on the same CPGs content. Among these models, the Deep Learning models achieved higher accuracy compared to other models. For the embedding layer of the deep learning model we experiment, the model accuracy without and with various word embedding techniques, including C\&W, CBOW, GloVE \cite{pennington2014glove}, Word2vec \cite{mikolov2013efficient}, and IWV \cite{rezaeinia2019sentiment}. The accuracy of the model differs by applying various embedding techniques. As shown in Figure \ref{fig:wordEmbeddingAccuracy}, the accuracy increased by 8\% when we used IWV in the embedding layer of the deep learning model.

\begin{figure}[H]
	\centering
	\includegraphics [scale=0.90]{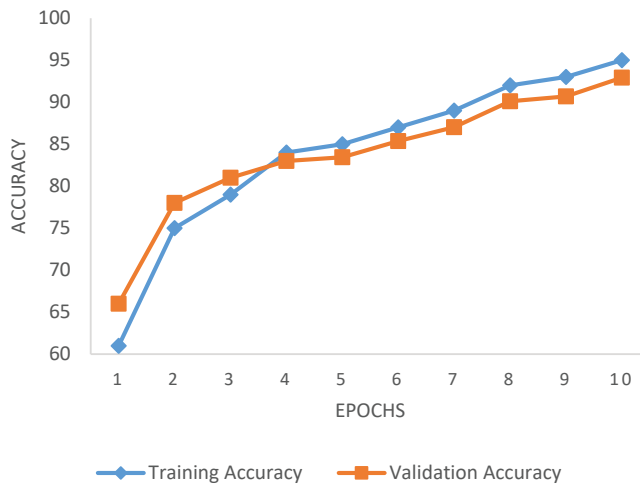}
	\caption{Model accuracy in various epochs.}
	\label{fig:epochsAccuracy}
\end{figure}

Therefore, we used the IWV based deep Learning model in the study for the classification of unseen CPG sentences because of the highest accuracy. We checked the model accuracy at different number of epochs, and it showed best accuracy at ten epochs, as shown in Figure \ref{fig:epochsAccuracy}. Therefore, we trained and validated it using ten epochs.

\subsection{Rules Generation }
\label{ResultsRulesGeneration}

We evaluated the rule generation process using user centric evaluation. The steps described earlier extracted twenty one rules from the CPG. To validate the methodology, we compared the result of our methodology with human experts extracted rules. We invited three physicians from our collaborative hospital, they have more than five years of experience in knowledge extraction from CPG. They extracted the rules in  \textbf{IF} \textit{Conditions} \textbf{THEN} \textit{Action} format. All the extracted rules were collected and manually processed to remove duplicates. We also aligned the diverse concepts, used by physicians, to uniform terminology. The list of rules extracted by the proposed methodology and human experts is shown in Table \ref{tab:extractedRules}. Where column $R_{we}$, $R_{exp1}$, $R_{exp2}$, and $R_{exp3}$ represent the rules extracted by our proposed methodology, expert 1, expert 2, and expert 3, respectively. 

\begin{small}
	\fontsize{8}{11}\selectfont
	\begin{longtable}{|c|p{200pt}|c|c|c|c|}
		\caption{Extracted rules by proposed method and experts} \label{tab:extractedRules} \\
		
		\hline \multicolumn{1}{|c|}{S.No} & \multicolumn{1}{c|}{Rules} & \multicolumn{1}{c|}{$R_{we}$} & \multicolumn{1}{c|}{$R_{exp1}$} & \multicolumn{1}{c|}{$R_{exp2}$} & \multicolumn{1}{c|}{$R_{exp3}$} \\ \hline 
		\endfirsthead
		
		\multicolumn{6}{c}%
		{{\bfseries \tablename\ \thetable{} -- continued from previous page}} \\
		\hline \multicolumn{1}{|c|}{S.No} & \multicolumn{1}{c|}{Rules} & \multicolumn{1}{c|}{$R_{we}$} & \multicolumn{1}{c|}{$R_{exp1}$} & \multicolumn{1}{c|}{$R_{exp2}$} & \multicolumn{1}{c|}{$R_{exp3}$} \\ \hline 
		\endhead
		
		\hline \multicolumn{6}{|r|}{{Continued on next page}} \\ \hline
		\endfoot
		
		\hline \hline
		\endlastfoot
		
		Rule 1 & \textbf{IF} Hypertension = Yes AND Age $\geq$ 60 \textbf{THEN} BP Goal $<$ 150/90 mm Hg. & \checkmark & \checkmark & \checkmark & \checkmark 	\\
		Rule 2 & \textbf{IF} Hypertension = Yes AND Age $<$ 60  \textbf{THEN} SBP Goal $<$ 140/90 mm Hg. & \checkmark & \checkmark & \checkmark & \texttimes \\
		Rule 3 &  \textbf{IF} Hypertension = Yes AND Age $<$ 60 AND Diabetes = Yes   \textbf{THEN} BP Goal $<$ 140/90 mm Hg. & \checkmark & \checkmark & \checkmark & \checkmark\\
		Rule 4 &  \textbf{IF} Hypertension = Yes  AND Diabetes = No and CKD = Yes   \textbf{THEN} BP Goal $<$ 140/90 mm Hg. & \texttimes & \checkmark & \texttimes & \checkmark\\
		Rule 5 &  \textbf{IF} Hypertension = Yes  AND Race = `Non-black'  \textbf{THEN} Drug treatment = angiotension-converting enzyme inhibitor, angiotensin receptor blokcer, calcium channel blocker, or thiazide-type diurectic. & \checkmark & \checkmark & \checkmark & \texttimes \\
		Rule 6 &  \textbf{IF} Hypertension = Yes  AND Race = `Non-black' and Diabetes = Yes  \textbf{THEN} Drug treatment = angiotension-converting enzyme inhibitor, angiotensin receptor blokcer, calcium channel blocker, or thiazide-type diurectic. & \texttimes & \checkmark & \checkmark & \checkmark \\
		Rule 7 &  \textbf{IF} Hypertension = Yes  AND Race = `black' \textbf{THEN} Initial Therapy = Calcium channel blocker OR Thizzide-type diurectic. & \checkmark & \texttimes & \checkmark & \texttimes \\
		Rule 8 &  \textbf{IF} Hypertension = Yes  AND Race = `black' AND Diabetes = Yes \textbf{THEN} Initial Therapy = Calcium channel blocker OR Thizzide-type diurectic. & \checkmark & \checkmark & \checkmark & \texttimes \\
		Rule 9 &  \textbf{IF} Hypertension = Yes  AND Age = 60 AND SBP $<$ 140 AND Adverse Effect = No \textbf{THEN} Continue Pharmacologic Treatment. & \texttimes & \texttimes & \texttimes & \checkmark\\		
		Rule 10 &  \textbf{IF} Hypertension = Yes  AND Race = `Non-black' \textbf{THEN} Initial Antihypertensive Treatment = Thiazide-type diurectic, calcium channel blocker, Angiotensin-converting enzyme inhibitor, or Angiotensin receptor blocker. & \checkmark & \checkmark & \checkmark & \checkmark\\
		Rule 11 &  \textbf{IF} Hypertension = Yes  AND Race = `Non-black' and Diabetes = Yes \textbf{THEN} Initial Antihypertensive Treatment = Thiazide-type diurectic, calcium channel blocker, Angiotensin-converting enzyme inhibitor, or Angiotensin receptor blocker. & \texttimes & \checkmark & \checkmark & \texttimes \\
		Rule 12 &  \textbf{IF} Hypertension = Yes  AND Race = `Black' \textbf{THEN}  Antihypertensive Treatment = Thiazide-type diurectic OR CCB. & \checkmark & \checkmark & \texttimes & \checkmark\\
		Rule 13 &  \textbf{IF} Hypertension = Yes  AND Race = `Black' AND Diabetes = Yes \textbf{THEN}  Antihypertensive Treatment = Thiazide-type diurectic OR CCB. & \checkmark & \texttimes & \checkmark & \texttimes \\
		Rule 14 &  \textbf{IF} Hypertension = Yes  AND Age = 18 AND CKD = Yes \textbf{THEN}  Antihypertensive Treatment = (-LRB- OR add-on-RRB-) and (ACEI OR ARB). & \checkmark & \checkmark & \checkmark & \checkmark\\
		Rule 15 &  \textbf{IF} Hypertension = Yes  AND Age = 18 AND CKD = Yes AND Diabetes = Yes \textbf{THEN}  Antihypertensive Treatment = (-LRB- OR add-on-RRB-) and (ACEI OR ARB). & \texttimes & \texttimes & \checkmark & \checkmark\\
		Rule 16 &  \textbf{IF} BP $>$ goal BP AND Treatment Duration = 1 Month  \textbf{THEN}  Increase does OR add drug from -LRB- Thiazide-type diurectic, CCB, ACEI, or ARB-RRB-. & \checkmark & \checkmark & \checkmark & \checkmark\\
		Rule 17 &  \textbf{IF} BP $>$ goal BP AND No of drugs = 2 \textbf{THEN}  Add third drug. & \texttimes & \checkmark & \texttimes & \texttimes \\
		Rule 18 &  \textbf{IF} Hypertension = Yes AND Race = `Black' \textbf{THEN}  Choose Thiazide-type diurentics Over ACEI. & \checkmark & \checkmark & \checkmark  & \checkmark \\
		Rule 19 &  \textbf{IF} Hypertension = Yes AND Race = `Black' \textbf{THEN}  First line therapy = Thiazide-type diurentics and CCBs. & \checkmark & \checkmark & \checkmark & \checkmark \\
		Rule 20 &  \textbf{IF} Hypertension = Yes AND Age Group = 'adult' AND Diabetes = Yes \textbf{THEN}  SBP goal $<$ 130 mm Hg. & \checkmark & \texttimes  & \texttimes & \texttimes \\
		Rule 21 &  \textbf{IF} Age $>$ 30 AND Age $<$ 60 AND DBP $>$ 90 \textbf{THEN}  Hypertension. & \texttimes & \texttimes & \checkmark & \texttimes \\
	\end{longtable}
\end{small}

We compared the proposed methodology extracted rules with the human experts extracted rules using Jaccard similarity Coefficient (JC). JC is a statistical measure used to evaluate similarities and differences between objects. The formula for calculating the JC is given in Equation \ref{eqn:jc}, where $R_{we}$ represents the rules extracted by our methodology, and $R_{exp}$ represents rules extracted by human experts. The values of JC lies between zero and one i.e $0 \leq JC(R_{we}, R_{exp}) \leq 1$, where $JC = 0$ represents that the rule extracted by our methodology is completely different from experts' extracted rule while $JC = 1$ represents that the extracted rules are equivalent. The comparison of the rules extracted via the proposed methodology and the experts extracted rules is given in Table \ref{tab:extractedRules}. Where  `\checkmark' represent the rule was extracted by the concern participant while `\texttimes' shows that it has missed the rule.

\begin{equation}
\label{eqn:jc}
JC(R_{we}, R_{exp}) = \dfrac{|R_{we} \cap R_{exp}|}{|R_{we}  \cup R_{exp}|}
\end{equation}

The JC values of the proposed methodology compare to expert1, expert2, and expert3 are 0.6, 0.7, and 0.4, respectively. The JC values indicate that the proposed methodology is feasible and reliable for transforming textual CPG to computable format.

\section{Conclusion}
\label{Conclusion}
This study introduced a methodology, which extracts knowledge rules from CPGs, automatically. It helps in transforming the latest research findings into clinical practice. The methodology mainly focuses on accurate classification of CPG sentences by utilizing deep learning techniques and extracts qualifiers to locate condition and action phrases in a sentence. The extracted phrases are processed to identify  conditions and corresponding actions and present it in the plain rule format. The methodology reduces the CPGs processing time, and the extracted knowledge can be part of clinical information system or clinical decision support system to assist evidence based standardized decision during real practices.

\section*{acknowledgements}
This research was supported by Institute for Information \& communications Technology Promotion(IITP) grant funded by the Korea government(MSIT)
(No.2017-0-00655).


\bibliographystyle{elsarticle-num}

\bibliography{mybibfile}

\begin{thebibliography}{10}
\expandafter\ifx\csname url\endcsname\relax
  \def\url#1{\texttt{#1}}\fi
\expandafter\ifx\csname urlprefix\endcsname\relax\def\urlprefix{URL }\fi
\expandafter\ifx\csname href\endcsname\relax
  \def\href#1#2{#2} \def\path#1{#1}\fi

\bibitem{kluegl2016uima}
P.~Kluegl, M.~Toepfer, P.-D. Beck, G.~Fette, F.~Puppe, Uima ruta: Rapid
  development of rule-based information extraction applications, Natural
  Language Engineering 22~(1) (2016) 1--40.

\bibitem{pandita2017tmap}
R.~Pandita, R.~Jetley, S.~Sudarsan, T.~Menzies, L.~Williams, Tmap: Discovering
  relevant api methods through text mining of api documentation, Journal of
  Software: Evolution and Process 29~(12) (2017) e1845.

\bibitem{pyon2013primer}
E.~Y. Pyon, Primer on clinical practice guidelines, Journal of pharmacy
  practice 26~(2) (2013) 103--111.

\bibitem{heins2016adherence}
M.~J. Heins, J.~D. de~Jong, I.~Spronk, V.~K. Ho, M.~Brink, J.~C. Korevaar,
  Adherence to cancer treatment guidelines: influence of general and
  cancer-specific guideline characteristics, The European Journal of Public
  Health 27~(4) (2016) 616--620.

\bibitem{ruan2015clinical}
X.~Ruan, L.~Ma, N.~Vo, S.~Chiravuri, Clinical practice guidelines: the more,
  the better?, North American Journal of Medicine and Science 8~(2) (2015).

\bibitem{upshur2014clinical}
R.~E. Upshur, Do clinical guidelines still make sense? no, The Annals of Family
  Medicine 12~(3) (2014) 202--203.

\bibitem{ryan2017adherence}
M.~A. Ryan, Adherence to clinical practice guidelines, Otolaryngology--Head and
  Neck Surgery 157~(4) (2017) 548--550.

\bibitem{wang2018clinical}
Y.~Wang, L.~Wang, M.~Rastegar-Mojarad, S.~Moon, F.~Shen, N.~Afzal, S.~Liu,
  Y.~Zeng, S.~Mehrabi, S.~Sohn, et~al., Clinical information extraction
  applications: a literature review, Journal of biomedical informatics 77
  (2018) 34--49.

\bibitem{sun2013evaluating}
W.~Sun, A.~Rumshisky, O.~Uzuner, Evaluating temporal relations in clinical
  text: 2012 i2b2 challenge, Journal of the American Medical Informatics
  Association 20~(5) (2013) 806--813.

\bibitem{sarawagi2008information}
S.~Sarawagi, et~al., Information extraction, Foundations and
  Trends{\textregistered} in Databases 1~(3) (2008) 261--377.

\bibitem{small2014review}
S.~G. Small, L.~Medsker, Review of information extraction technologies and
  applications, Neural computing and applications 25~(3-4) (2014) 533--548.

\bibitem{rezaeinia2019sentiment}
S.~M. Rezaeinia, R.~Rahmani, A.~Ghodsi, H.~Veisi, Sentiment analysis based on
  improved pre-trained word embeddings, Expert Systems with Applications 117
  (2019) 139--147.

\bibitem{hettne2010rewriting}
K.~M. Hettne, E.~M. van Mulligen, M.~J. Schuemie, B.~J. Schijvenaars, J.~A.
  Kors, Rewriting and suppressing umls terms for improved biomedical term
  identification, Journal of biomedical semantics 1~(1) (2010) 5.

\bibitem{jacobsen2009clinical}
P.~B. Jacobsen, Clinical practice guidelines for the psychosocial care of
  cancer survivors: current status and future prospects, Cancer 115~(S18)
  (2009) 4419--4429.

\bibitem{zhao2019associative}
S.~Zhao, L.~Li, H.~Lu, A.~Zhou, S.~Qian, Associative attention networks for
  temporal relation extraction from electronic health records, Journal of
  Biomedical Informatics 99 (2019) 103309.

\bibitem{kim2019ensemble}
Y.~Kim, S.~M. Meystre, Ensemble method--based extraction of medication and
  related information from clinical texts, Journal of the American Medical
  Informatics Association 27~(1) (2019) 31--38.

\bibitem{viegas2019cluwords}
F.~Viegas, S.~Canuto, C.~Gomes, W.~Luiz, T.~Rosa, S.~Ribas, L.~Rocha, M.~A.
  Gon{\c{c}}alves, Cluwords: Exploiting semantic word clustering representation
  for enhanced topic modeling, in: Proceedings of the Twelfth ACM International
  Conference on Web Search and Data Mining, ACM, 2019, pp. 753--761.

\bibitem{zheng2017effective}
S.~Zheng, J.~J. Lu, N.~Ghasemzadeh, S.~S. Hayek, A.~A. Quyyumi, F.~Wang,
  Effective information extraction framework for heterogeneous clinical reports
  using online machine learning and controlled vocabularies, JMIR medical
  informatics 5~(2) (2017) e12.

\bibitem{i2b2}
i2b2, {Informatics for Integrating Biology and the Bedside},
  \url{https://www.i2b2.org/}, [Online; accessed 01-March-2020].

\bibitem{i2b2Challenges}
i2b2, {Challenges},
  \url{https://www.i2b2.org/NLP/RDoCforPsychiatry/PreviousChallenges.php/},
  [Online; accessed 01-March-2020].

\bibitem{n2c2}
n2c2, {National NLP Clinical Challenges},
  \url{https://n2c2.dbmi.hms.harvard.edu/}, [Online; accessed 01-March-2020].

\bibitem{chapman2011overcoming}
W.~W. Chapman, P.~M. Nadkarni, L.~Hirschman, L.~W. D'avolio, G.~K. Savova,
  O.~Uzuner, Overcoming barriers to nlp for clinical text: the role of shared
  tasks and the need for additional creative solutions (2011).

\bibitem{kreimeyer2017natural}
K.~Kreimeyer, M.~Foster, A.~Pandey, N.~Arya, G.~Halford, S.~F. Jones,
  R.~Forshee, M.~Walderhaug, T.~Botsis, Natural language processing systems for
  capturing and standardizing unstructured clinical information: a systematic
  review, Journal of biomedical informatics 73 (2017) 14--29.

\bibitem{alves2019information}
S.~Alves, J.~Costa, J.~Bernardino, Information extraction applications for
  clinical trials: A survey, in: 2019 14th Iberian Conference on Information
  Systems and Technologies (CISTI), IEEE, 2019, pp. 1--6.

\bibitem{sun2018data}
W.~Sun, Z.~Cai, Y.~Li, F.~Liu, S.~Fang, G.~Wang, Data processing and text
  mining technologies on electronic medical records: a review, Journal of
  healthcare engineering 2018 (2018).

\bibitem{zhang2019deep}
T.~Zhang, J.~Leng, Y.~Liu, Deep learning for drug--drug interaction extraction
  from the literature: a review, Briefings in bioinformatics (2019).

\bibitem{peleg2003comparing}
M.~Peleg, S.~Tu, J.~Bury, P.~Ciccarese, J.~Fox, R.~A. Greenes, R.~Hall, P.~D.
  Johnson, N.~Jones, A.~Kumar, et~al., Comparing computer-interpretable
  guideline models: a case-study approach, Journal of the American Medical
  Informatics Association 10~(1) (2003) 52--68.

\bibitem{bottrighi2009analysis}
A.~Bottrighi, F.~Chesani, P.~Mello, M.~Montali, S.~Montani, S.~Storari,
  P.~Terenziani, Analysis of the glare and gprove approaches to clinical
  guidelines, in: International Workshop on Knowledge Representation for Health
  Care, Springer, 2009, pp. 76--87.

\bibitem{ten2008computer}
A.~Ten~Teije, S.~Miksch, P.~Lucas, Computer-based medical guidelines and
  protocols: a primer and current trends, Vol. 139, Ios Press, 2008.

\bibitem{latoszek2010clinical}
A.~Latoszek-Berendsen, H.~Tange, H.~Van Den~Herik, A.~Hasman, From clinical
  practice guidelines to computer-interpretable guidelines, Methods of
  information in medicine 49~(06) (2010) 550--570.

\bibitem{qiu2018automated}
Y.~Qiu, P.~Tang, H.~Wang, J.~Zhang, X.~Qin, Automated encoding of clinical
  guidelines into computer-interpretable format, in: Proceedings of the 2018
  6th International Conference on Bioinformatics and Computational Biology,
  ACM, 2018, pp. 138--144.

\bibitem{khodambashi2017reviewing}
S.~Khodambashi, {\O}.~Nytr{\o}, Reviewing clinical guideline development tools:
  features and characteristics, BMC medical informatics and decision making
  17~(1) (2017) 132.

\bibitem{de2004approaches}
P.~A. De~Clercq, J.~A. Blom, H.~H. Korsten, A.~Hasman, Approaches for creating
  computer-interpretable guidelines that facilitate decision support,
  Artificial intelligence in medicine 31~(1) (2004) 1--27.

\bibitem{shahar1998asgaard}
Y.~Shahar, S.~Miksch, P.~Johnson, The asgaard project: a task-specific
  framework for the application and critiquing of time-oriented clinical
  guidelines, Artificial intelligence in medicine 14~(1-2) (1998) 29--51.

\bibitem{shiffman2000gem}
R.~N. Shiffman, B.~T. Karras, A.~Agrawal, R.~Chen, L.~Marenco, S.~Nath, Gem: a
  proposal for a more comprehensive guideline document model using xml, Journal
  of the American Medical Informatics Association 7~(5) (2000) 488--498.

\bibitem{ohno1998guideline}
L.~Ohno-Machado, J.~H. Gennari, S.~N. Murphy, N.~L. Jain, S.~W. Tu, D.~E.
  Oliver, E.~Pattison-Gordon, R.~A. Greenes, E.~H. Shortliffe, G.~O. Barnett,
  The guideline interchange format: a model for representing guidelines,
  Journal of the American Medical Informatics Association 5~(4) (1998)
  357--372.

\bibitem{kim2008modeling}
S.~Kim, P.~J. Haug, R.~A. Rocha, I.~Choi, Modeling the arden syntax for medical
  decisions in xml, International journal of medical informatics 77~(10) (2008)
  650--656.

\bibitem{sutton2003syntax}
D.~R. Sutton, J.~Fox, The syntax and semantics of the pro forma guideline
  modeling language, Journal of the American Medical Informatics Association
  10~(5) (2003) 433--443.

\bibitem{tu2007sage}
S.~W. Tu, J.~R. Campbell, J.~Glasgow, M.~A. Nyman, R.~McClure, J.~McClay,
  C.~Parker, K.~M. Hrabak, D.~Berg, T.~Weida, et~al., The sage guideline model:
  achievements and overview, Journal of the American Medical Informatics
  Association 14~(5) (2007) 589--598.

\bibitem{james20142014}
P.~A. James, S.~Oparil, B.~L. Carter, W.~C. Cushman, C.~Dennison-Himmelfarb,
  J.~Handler, D.~T. Lackland, M.~L. LeFevre, T.~D. MacKenzie, O.~Ogedegbe,
  et~al., 2014 evidence-based guideline for the management of high blood
  pressure in adults: report from the panel members appointed to the eighth
  joint national committee (jnc 8), Jama 311~(5) (2014) 507--520.

\bibitem{chow2012idsa}
A.~W. Chow, M.~S. Benninger, I.~Brook, J.~L. Brozek, E.~J. Goldstein, L.~A.
  Hicks, G.~A. Pankey, M.~Seleznick, G.~Volturo, E.~R. Wald, et~al., Idsa
  clinical practice guideline for acute bacterial rhinosinusitis in children
  and adults, Clinical Infectious Diseases 54~(8) (2012) e72--e112.

\bibitem{british2003scottish}
B.~T. Society, Scottish intercollegiate guidelines network, British Guideline
  on the management of asthma. Thorax 58~(Suppl 1) (2003) i1--94.

\bibitem{mikolov2013efficient}
T.~Mikolov, K.~Chen, G.~Corrado, J.~Dean, Efficient estimation of word
  representations in vector space, arXiv preprint arXiv:1301.3781 (2013).

\bibitem{pennington2014glove}
J.~Pennington, R.~Socher, C.~Manning, Glove: Global vectors for word
  representation, in: Proceedings of the 2014 conference on empirical methods
  in natural language processing (EMNLP), 2014, pp. 1532--1543.

\bibitem{hussain2018multimodal}
J.~Hussain, W.~A. Khan, T.~Hur, H.~S.~M. Bilal, J.~Bang, A.~U. Hassan,
  M.~Afzal, S.~Lee, A multimodal deep log-based user experience (ux) platform
  for ux evaluation, Sensors 18~(5) (2018) 1622.

\bibitem{mierswa2006yale}
I.~Mierswa, M.~Wurst, R.~Klinkenberg, M.~Scholz, T.~Euler, Yale: Rapid
  prototyping for complex data mining tasks, in: Proceedings of the 12th ACM
  SIGKDD international conference on Knowledge discovery and data mining, ACM,
  2006, pp. 935--940.

\bibitem{abadi2016tensorflow}
M.~Abadi, A.~Agarwal, P.~Barham, E.~Brevdo, Z.~Chen, C.~Citro, G.~S. Corrado,
  A.~Davis, J.~Dean, M.~Devin, et~al., Tensorflow: Large-scale machine learning
  on heterogeneous distributed systems, arXiv preprint arXiv:1603.04467 (2016).

\end{thebibliography}

\end{document}